\documentclass{article} 
\usepackage{ICLR23/iclr2023_conference,times}

\usepackage{amsmath,amsfonts,bm}

\def\eqref#1{equation~\ref{#1}}

\def\1{\bm{1}}

\DeclareMathAlphabet{\mathsfit}{\encodingdefault}{\sfdefault}{m}{sl}
\SetMathAlphabet{\mathsfit}{bold}{\encodingdefault}{\sfdefault}{bx}{n}

\usepackage{hyperref}
\usepackage{url}
\usepackage{booktabs}
\usepackage{multicol}
\usepackage{multirow}
\usepackage{comment}
\usepackage{pgfplots}
\usepackage{subcaption}
\usepackage{microtype}
\usepackage{arydshln}

\definecolor{forestgreen}{HTML}{009B55}
\definecolor{sepia}{HTML}{671800}
\definecolor{midnightblue}{HTML}{006795}
\definecolor{orangered}{HTML}{E24C00}
\definecolor{bblue}{HTML}{4F81BD}
\definecolor{rred}{HTML}{C0504D}
\definecolor{ggreen}{HTML}{9BBB59}
\definecolor{ppurple}{HTML}{9F4C7C}

\title{Query Refinement Prompts for Closed-Book Long-Form Question Answering}

\author{Reinald Kim Amplayo \\ \textbf{Kellie Webster} \\ \textbf{Michael Collins} \\ \textbf{Dipanjan Das} \\ \textbf{Shashi Narayan} \\ 
Google Research \\ 
\texttt{\small \{reinald, websterk, mjcollins, dipanjand, shashinarayan\}@google.com}
}

\author{Reinald Kim Amplayo \\ Google Research \\ \texttt{\small reinald@google.com} \And 
Kellie Webster \\ Google Research \\ \texttt{\small websterk@google.com} \And 
Michael Collins \\ Google Research \\ \texttt{\small mjcollins@google.com} \And
Dipanjan Das \\ Google Research \\ \texttt{\small dipanjand@google.com} \And 
Shashi Narayan \\ Google Research \\ \texttt{\small shashinarayan@google.com}
\And \And \And ~~~~~~~ 
}

\iclrfinalcopy
\begin{document}
\maketitle
\begin{abstract}
Large language models (LLMs) have been shown to perform well in answering questions and in producing long-form texts, both in few-shot closed-book settings.
While the former can be validated using well-known evaluation metrics, the latter is difficult to evaluate.
We resolve the difficulties to evaluate long-form output by doing
both tasks at once -- to do question answering that requires long-form answers.
Such questions tend to be \textit{multifaceted}, i.e., they may have ambiguities and/or require information from multiple sources. To this end, we define {\em query refinement} prompts that encourage LLMs to explicitly express the multifacetedness in questions and generate long-form answers covering 
multiple facets of the question.
Our experiments on two long-form question answering datasets, ASQA and AQuAMuSe, show that using our prompts allows us to outperform fully finetuned models in the closed book setting, as well as achieve results comparable to retrieve-then-generate open-book models.
\end{abstract}

\section{Introduction}

Interest in large language models (LLMs) has exploded in recent years due to their wide range of abilities improving the state-of-the-art with just a few or no examples \citep{brown2020language,wei2022finetuned,chowdhery2022palm,wei2022chain}.
One task that benefited greatly is closed-book question answering \citep{roberts-etal-2020-much} in a few-shot setting, i.e., to produce correct answers to questions without access to passages to read and find the answer.
Most of the impressive results, however, are limited to generating short answers, and while previous work has utilized LLMs to generate long-form texts \citep{elkins2020can,wei2022chain}, most of these outputs are difficult to evaluate since they are subjective in nature (i.e., they may have multiple correct and distinctive answers).
In this paper, we attempt to evaluate the long-form generation ability of LLMs through long-form question answering \citep{fan-etal-2019-eli5} in the closed-book setting.

Unlike question answering with short answers, questions with long-form answers are naturally multifacted, covering many aspects that are required to fully answer the questions (see Table~\ref{tab:example} for examples). This introduces several challenges that LLMs would need to resolve:
\begin{enumerate}
    \item{\textbf{Ambiguity}}: Many questions in a real-world setting are ambiguous in some way; \citet{min-etal-2020-ambigqa} estimated that more than half of Natural Questions \citep{kwiatkowski-etal-2019-natural} have multiple plausible interpretations. In the first example of Table~\ref{tab:example}, the question is ambiguous owing to there being multiple types of cars. A system producing a long-form answer to this question should present the multiple answers in a coherent natural language text.
    \item{\textbf{Information Consolidation}}: Some questions require consolidating information from multiple sources to be able to fully answer them \citep{kulkarni2020aquamuse}. In the second example of Table~\ref{tab:example}, information regarding the British and Dutch origins of the term may be found in two different documents and LLMs should therefore be able to synthesize knowledge from multiple sources in a coherent manner.
    \item{\textbf{Correctness}}: Having a model to answer questions in a closed-book setting requires it to leverage content it has learned during pretraining. While previous work \citep{roberts-etal-2020-much,brown2020language} has shown that LLMs can do so to produce factoid answers, generating long-form answers requires more and complex information to be leveraged.
\end{enumerate}

\begin{table}
    \centering
    \footnotesize
    \begin{tabular}{p{0.12\textwidth}p{0.63\textwidth}p{0.16\textwidth}}
    \toprule
    \textbf{ASQA ~Question} & \multicolumn{2}{c}{\texttt{when was the first car made in america?}} \\
    \midrule
    \multirow{2}{0.1\textwidth}{\textbf{Multiple Facets}} & When was the first carriage-sized automobile made in America? & 1871 \\
       & When was the first running, gasoline-powered car made in America? & 1893 \\
    & ... & ... \\
    \midrule
    \textbf{Long-form Answer} & \multicolumn{2}{p{0.82\textwidth}}{\textit{The first carriage-sized automobile that could be driven on wagon roads in the United States was steam-powered and invented in 1871 by Dr. J.W. Carhart in Racine, Wisconsin. The first running, gasoline-powered car that was made in America, the Duryea Motor Wagon, was built in 1893. The Studebaker Automobile Company, which started building cars in 1897, sold electric vehicles in 1902, and gasoline vehicles in 1904.}} \\
    \midrule
    \textbf{AQuAMuSe Question} & \multicolumn{2}{c}{\texttt{where did the term shooting brake come from?}} \\
    \midrule
    \multirow{2}{0.1\textwidth}{\textbf{Multiple Facets}} 
       & When did the term originate? & early 19th century \\
       & As what term did the term originate? & as a British term \\
    & ... & ... \\
    \midrule
    \textbf{Long-form Answer} & \multicolumn{2}{p{0.82\textwidth}}{\textit{``Shooting-brake'' originated as an early 19th century British term for a vehicle used to carry shooting parties with their equipment and game. The etymology of the term brake is uncertain; initially a chassis used to break in horses, and subsequently used to describe a motorized vehicle. It is also possible, that the word `brake' has its origins in the Dutch word `brik' which means `cart' or `carriage'.}} \\
    \bottomrule
    \end{tabular}
    \caption{Example multifaceted questions from ASQA and AQuAMuSe and their corresponding answers. A system for closed-book question answering needs to understand that some questions have multiple valid answers and synthesise these into a coherent natural language text output.}
    \label{tab:example}
\end{table}

Our study is based on two long-form question answering benchmarks, ASQA \citep{stelmakh2022asqa} and AQuAMuSe \citep{kulkarni2020aquamuse}, which focus on queries that may be ambiguous and/or require information from multiple sources.
Using these two datasets, our work is the first to show that LLMs are capable of generating long-form answers to complex questions of various types in a closed-book setting. We devise novel {\em query refinement} prompts that encourage LLMs to express multiple facets of input question and generate multifaceted answers discussing all identified interpretations coherently. 
Specifically, we identify several different types of multifacetedness in questions and produce a labeled set of query refinement prompts for question answer pairs with a balanced coverage over different types. 
We then introduce an intermediate query refinement step, a generation subtask akin to explanations and reasoning chains \citep{wei2022chain,zhou2022least}, where the goal is to identify multiple facets of a given question.

We evaluate our long-form answers using ROUGE (measuring stylistic similarity to gold answers; \citeauthor{lin-2004-rouge}, \citeyear{lin-2004-rouge}) and reading comprehension models (measuring correctness; \citeauthor{stelmakh2022asqa}, \citeyear{stelmakh2022asqa}). 
Using our query refinement prompts in the few-shot prompting and prompt tuning \citep{lester-etal-2021-power} settings, we are able to achieve significantly better performance to fully finetuned closed-book systems, as well as comparable performance to open-book retrieve-then-generate systems, both finetuned on the full training dataset.

\section{Related Work}

\subsection{Prompting in Large Language Models}

Large language models have the revolutionary ability to generalize to tasks presented with natural language prompts \citep{brown2020language,chowdhery2022palm}. This ability has been mostly attributed to the large scale of these models and the learning objective of predicting the next token \citep{brown2020language}. 
The prompting ability of LLMs has been used successfully in text classification, reading comprehension and open-domain question answering tasks.
It has also been shown that LLMs improve on complex reasoning tasks by generating intermediate long-form texts in the form of explanations or reasoning steps \citep{wei2022chain,zhou2022least}.
However, prompting is still ineffective when tasked to generate long-form outputs as an \textit{end task}, e.g., generating long summaries for summarization \citep{brown2020language,chowdhery2022palm}.
Our work is the first to show that LLMs can do long-form text generation through question answering with the help of a refinement step in the prompt.


Our work is related to and inspired by work on reasoning chains in LLMs \citep{wei2022chain,zhou2022least,kojima2022large,snell2022learning}, where the goal is to explicitly generate a reasoning or an explanation before producing an answer. Most of these papers focus on arithmetic and commonsense reasoning questions, where reasoning and explanations are obvious. In this paper, we show that such intermediate explanation generation can also be helpful on tasks that implicitly involve multiple steps, such as long-form question answering where question refinement is necessary.
Moreover, we are the first to explore structured explanations in the form of a list of answer facets, which is shown in our experiments to be more effective than natural language explanations.


Prompting is just one way of using the LLMs.
There are several work \citep{multitask_prompted_training,flan,chowdhery2022palm} that attempted to finetune LLMs entirely for text generation tasks, which can be very expensive.
Prompt tuning \citep{lester-etal-2021-power} is a popular alternative where \textit{soft} prompts are prepended into the input and are finetuned. 
There are several other alternatives to prompting that show promising results for generating long-form texts, such as prefix tuning \citep{li-liang-2021-prefix}, adapters \citep{bapna-firat-2019-simple}, and several parameter-efficient finetuning techniques \citep{clive2021control,he2022towards,liu2022psp} that introduce new parameters to the model that is updated during training while leaving the LLM parameter fixed.
All of these, however, still require several forward passes to the LLMs, which may not be available in many cases.
Nevertheless, we show that applying our query refinement step in prompt tuning improves the performance. 

\subsection{Long-form Question Answering}

Question answering has emerged as a key way to discover and demonstrate advances in large language models, which are showing their skill on increasingly difficult formulations of the task.
SQuAD \citep{rajpurkar-etal-2016-squad} proposed the first large-scale, human-created reading comprehension task and was used to show the promise of neural architectures, which quickly attained human-like performance on the dataset.
Since, there has been a proliferation of reading comprehension datasets developed which probe for specific capabilities \citep{joshi-etal-2017-triviaqa,choi-etal-2018-quac,reddy-etal-2019-coqa,quizbowl}.
The Natural Questions \citep{kwiatkowski-etal-2019-natural} effort provided a large reading comprehension dataset based on real information-seeking queries to the Google search engine, and has served most recently a basis for the exploration of questions where a simple short answer is not sufficient to address the information need of a complex question.

One response strategy to such questions is a long-form answer, studied here. Both ASQA \citep{stelmakh2022asqa} and AQuaMuSe \citep{kulkarni2020aquamuse} require that systems consolidate information from multiple sources to generate multifaceted long-form answers to questions from the Natural Questions.
ASQA focuses on the subset of questions labeled in AmbigQA \citep{min-etal-2020-ambigqa} for which it is possible to enumerate a collection of refinements and factoid answers that should be covered in a long-form answer. 
On the other hand, AQuaMuSe focuses on questions without short factoid answers, that typically have a looser relationship to one another.
We study both so as to understand what prompting strategies work for the different style of reasoning required to do well on each.
ELI5 \citep{fan-etal-2019-eli5} is another long-form question answering dataset that was automatically gathered from Reddit threads, but subsequent work \citep{krishna-etal-2021-hurdles} has shown problems in its evaluation, including training/validation overlap and gameable metrics.

Finally, our work is also related to query-focused summarization \citep{dang2005overview,zhong-etal-2021-qmsum,kulkarni2020aquamuse}, where a set of relevant passages is assumed to be available. AQuaMuSe was developed for this task but, where our experiments are in the closed-book setting, we discard the given passages.

\section{Closed-book Long-form Question Answering}

\subsection{Few-Shot Prompting Formulation}

Given a question $q$, the goal of closed-book long-form question answering is to produce a passage-length text $a$ without access to external context or knowledge (beyond what was seen in pretraining). For the related task of closed-book \textit{factoid} question answering \citep{rajpurkar-etal-2016-squad,joshi-etal-2017-triviaqa}, this can been achieved with large language models using a \textit{few-shot prompting} setup \citep{brown2020language}. That is, given $k$ in-context exemplars of question-answer pairs $[(q'_1, a'_1),...,(q'_k, a'_k)]$, usually preceded by an instruction, an LLM will output an answer $a$ for question $q$ from knowledge stored in its parameters \citep{roberts-etal-2020-much}.


When the answer is instead long-form (see examples in Table~\ref{tab:example}), there are three subtasks that the model need to do to produce an answer: (1) Determining multiple facets of the question (\textit{Facet Identification}), (2) Retrieving multiple answers to the multiple facets of the question (\textit{Multifaceted Question Answering}), and (3) Realizing a long-form text that includes the multiple answers in a coherent manner (\textit{Surface Realization with Information Consolidation}). 

In the next sections, we extend the standard few-shot prompt in three different ways to help LLMs explicitly do these steps before arriving to an answer.
Firstly, we identify several types of multifacetedness in questions and produce a labeled and balanced set of exemplars (Section~\ref{sec:qtypes}). Next, we introduce a query refinement step in few-shot prompting that instructs the model to explicitly do the intermediate subtasks  (Section~\ref{sec:explanations}). Finally, we dynamically select exemplars to form a $k$-shot prompt based on similarity (Section~\ref{sec:dynamic}). 

\subsection{Types of Multifaceted Questions}
\label{sec:qtypes}

There are multiple reasons why a long-form answer would be more felicitous than a factoid answer to a question. Table~\ref{tab:qtypes} shows six common types of multifaceted questions in the ASQA \citep{stelmakh2022asqa} and AQuAMuSe datasets that fall into this category. These are highly related to the ambiguity categories in AmbigQA \citep{min-etal-2020-ambigqa}, which we used as a seed set for exploring the data in this work. To encourage further work in multifaceted question categorization, we detail the criteria we used to determine what type an example demonstrates. We produce a manually labeled set that contains 20 exemplars of each type to form a pool of possible training instances (see Table~\ref{tab:qtypes} for examples). When an example exhibited multiple types simultaneously, we opted not to include it as an exemplar.

\begin{table}[t!]
    \centering
    \footnotesize
    \begin{tabular}{p{13.5cm}}
        \toprule
        (a) \textbf{Conditional}: \textit{The original question needs to be refined by specifying additional} conditions \textit{that may be specifications or constraints.} \\
        \textit{Question}: When did movies start being made in color? \\
        \textit{Multifaceted QA Pairs}: \\
        \quad Q: When was the first film made that utilized any type of color? A: September 1, 1902 \\
        \quad Q: When did the first feature length film come out that was made entirely in three-strip Technicolor? A: June 13, 1935 \\
        Related AmbigQA categories: Event references, Properties \\
        \midrule
        (b) \textbf{Set-Valued}: \textit{The answer to the question is a unstructured collection of size two or greater.} \\
        \textit{Question}: What are the neighboring countries of South Korea? \\
        \textit{Multifaceted QA Pairs}: \\
        \quad Q: What are the neighboring countries to the North of South Korea? A: North Korea \\
        \quad Q: What are the neighboring countries to the South of South Korea? A: Japan \\
        Related AmbigQA category: $\emptyset$ \\
        \midrule
        (c) \textbf{Time Dependent}: \textit{The answer depends on the time at which the question was asked, or changed over time in the past.} \\
        \textit{Question}: Where was indian independence league formed in 1942? \\
        \textit{Multifaceted QA Pairs}: \\
        \quad Q: Where was indian independence league brought together in March 1942? A: Tokyo \\
        \quad Q: Where was indian independence league brought together in June 1942? A: Bangkok Conference \\
        Related AmbigQA category: Time-dependency\\
        \midrule
        (d) \textbf{Underspecified Reference}: \textit{There is a noun phrase in the question which may be resolved in multiple ways.} \\
        \textit{Question}: When did bat out of hell come out? \\
        \textit{Multifaceted QA Pairs}: \\
        \quad Q: When did the album bat out of hell come out? A: October 21, 1977 \\
        \quad Q: When did the TV series bat out of hell come out? A: 26 November 1966 \\
        Related AmbigQA category: Entity references \\
        \midrule
        (e) \textbf{Underspecified Type}: \textit{The entity type or sub-type is not specified in the question.} \\
        \textit{Question}: Who is the mayor in horton hears a who? \\
        \textit{Multifaceted QA Pairs}: \\
        \quad Q: Who plays the mayor in the 2008 film Horton Hears a Who? A: Steve Carell \\
        \quad Q: Who is the mayor in the 2008 film Horton Hears a Who? A: Mayor Ned McDodd \\
        Related AmbigQA category: Answer types \\
        \midrule
        (f) \textbf{Needs Elaboration}: \textit{The answer needs to be elaborated to fully answer the question} \\
        \textit{Question}: Where did ``you can't have your cake and eat it too'' come from? \\
        \textit{Multifaceted QA Pairs}: \\
        \quad Q: Where was the early recording of the phrase found? A: in a letter on 14 March 1538 \\
        \quad Q: Who sent the letter? A: Thomas, Duke of Norfolk \\
        \quad Q: To whom was the letter sent to? A: Thomas Cromwell \\
        \quad Q: How was it phrased in the letter: A: ``a man cannot have his cake and eat his cake'' \\
        Related AmbigQA category: $\emptyset$ \\
        \bottomrule
    \end{tabular}
    \caption{Six types of multifacetedness in questions. The first five types are sampled from the ASQA dataset, while the last type is sampled from the AQuAMuSe dataset. For each type, we also include the related AmbigQA \citep{min-etal-2020-ambigqa} categories if there are any.}
    \label{tab:qtypes}
\end{table}

\subsection{Query Refinement Step}
\label{sec:explanations}

Given that long-form question answering is essentially sequentially solving three subtasks, we propose to use an intermediate step that splits the tasks into two parts. This forces the LLM to explicitly resolve the intermediate subtasks before producing a long-form answer. We experimented with three types of refinements:


\begin{itemize}
    \item \textbf{Natural Language Explanations (NL)}: A sentence that explains why the question is ambiguous or needs elaboration. This refinement step essentially does \textit{Facet Identification}, i.e., it identifies the multiple facets of a question, which can either be disambiguations of the question, or things that need to be elaborated to fully answer the question.
    In previous work, LLMs have been used to generate NL explanations, e.g. for commonsense reasoning \citep{ji-etal-2020-generating} and jokes \citep{chowdhery2022palm}.
    \item \textbf{Lists of QA Pairs (QA)}: A list of question-answer pairs. We use the multifaceted QA pairs as shown in Table~\ref{tab:qtypes}. 
    This refinement does both \textit{Facet Identification} and \textit{Multifaceted Question Answering}, which means that when producing the answer, the model just needs to consolidate information into a coherent long-form answer.
    This is related to question answering blueprints \citep{narayan2022conditional}, which are used as content plans for summarization models, but their usage has not yet been explored in LLMs.
    \item \textbf{Lists of Answer Facets (AF)}: A list of \textit{answer facets}, pairings of the multiple facets of the question and their corresponding answer/s in the form ``\texttt{Facet}: \texttt{Answer}''. This is a concise version of the BP refinement, where we remove uninformative tokens such as wh-words and those that are repeated in the original question.
\end{itemize}

Table~\ref{tab:refinements} shows all three refinement strategies for all identified ambiguity types. The refinement is inserted between the question and the answer in the exemplar. At inference time, the LLM needs to generate both the refinement and the long-form answer. An example prompt with a query refinement step is illustrated in Figure~\ref{fig:asqa-prompt-template} of Appendix~\ref{sec:prompt-templates}.

\begin{table}[t!]
    \centering
    \footnotesize
    \begin{tabular}{p{13.5cm}}
        \toprule
        \textit{Question}: When did movies start being made in color? \\
        \midrule
        \textit{QA Refinement}: \\
        \quad Q: When was the first film made that utilized any type of color? A: September 1, 1902 \\
        \quad Q: When did the first feature length film come out that was made entirely in three-strip Technicolor? A: June 13, 1935 \\
        \textit{NL Refinement}: The answer depends on what is meant by in color (any type of color or three-strip Technicolor). \\
        \textit{AF Refinement}: \\
        \quad - any type of color: September 1, 1902 \\
        \quad - three-strip Technicolor: June 13, 1935 \\
        \midrule
        \textit{Answer}: The first film that utilized any type of color was made September 1, 1902. Edward Raymond Turner shot test footage to demonstrate his system, but projecting it proved problematic and he died a year later without having satisfactorily projected the footage. Later, in 2012, curators at the National Media Museum in Bradford, UK, were able to take the original custom-format nitrate film copied to black-and-white 35 mm film, which was then scanned into a digital video format by telecine and digital image processing was used to align and combine each group of three frames into one color image so that these films from 1902 could become viewable in full color. The first feature length film that was made entirely in three-strip Technicolor, which was an early process where color filters were used to photograph the color components as completely separate images, came out on June 13, 1935. \\
        \bottomrule
    \end{tabular}
    \caption{An example exemplar with QA, NL, and AF refinements. One exemplar in the $k$-shot prompt would include a question, one of the three refinements, and the answer. At inference time, the model would need to generate both the refinement and the answer.}
    \label{tab:refinements}
\end{table}

\subsection{Dynamic Prompting}
\label{sec:dynamic}

Finally, we form a k-shot prompt by selecting from our pool of exemplars created in Section~\ref{sec:qtypes}.
We do so using dynamic prompting (DP; \citealp{rubin-etal-2022-learning}), i.e., ranking exemplars $[(\hat{q}_1,\hat{a}_1),...,(\hat{q}_e,\hat{a}_e)]$ using the similarity between input question $q$ and candidate exemplar question $\hat{q}$. 
Dynamic prompting helps the model generate refinements for $q$ similarly to how refinements are done for a similar question $\hat{q}$.
We use BERTScore \citep{Zhang*2020BERTScore:} as our similarity metric.
The $k$ most similar exemplars are written to the prompt in reverse order, such that the most similar exemplar is written closest to input question $q$. 
In our experiments, we primarily used exemplars with questions labeled as ambiguous for ASQA (top 5 in Table~\ref{tab:qtypes}, for a total of 100 exemplars), and those that are labeled as \textit{Needs Elaboration} for AQuAMuSe (the last type in Table~\ref{tab:qtypes}, for a total of 20 exemplars). We also experimented combining both ASQA and AQuAMuSe exemplars in Section~\ref{sec:results-aquamuse}.



\section{Experiments on Ambiguous Question Answering}
\label{sec:results-asqa}

We conducted experiments on two question answering tasks that require long-form answers: Ambiguous Question Answering (ASQA; \citeauthor{stelmakh2022asqa}, \citeyear{stelmakh2022asqa}) and Query-focused Multi-document Summarization (AQuAMuSe; \citeauthor{kulkarni2020aquamuse}, \citeyear{kulkarni2020aquamuse}). 
In this section, we present results on ASQA; the following section presents results on AquaMuSe.

\subsection{ASQA Dataset}

The ASQA dataset \citep{stelmakh2022asqa} is a long-form question answering dataset built on top of the subset of ambiguous questions identified in the AmbigQA dataset \citep{min-etal-2020-ambigqa}, which itself is a subset of the NQ dataset \citep{kwiatkowski-etal-2019-natural}. ASQA consists of 4,353, 948, and 1,015 training, development, and test examples. 
Each question (e.g., \textit{Who directed Scarface?}) is paired with a list of QA pairs which signify disambiguated questions and their corresponding answers (e.g., \textit{Q: Who directed the 1932 film Scarface? A: Howard Hawks, Q: Who directed the 1983 film Scarface? A: Brian de Palma}). Finally, each example also has two human-written long-form answers based on the given the disambiguated QA pairs. Note that the list of disambiguated questions is not given at inference time.


\subsection{Evaluation}
Following \citet{stelmakh2022asqa}, we compare systems using three metrics:

\begin{itemize}
    \item \textbf{ROUGE-L} \citep{lin-2004-rouge}: Measures the comprehensibility of the system-generated answer with respect to the gold answers. Since there are two gold answers, we take the maximum ROUGE-L. We lowercase system and gold answers, and report on ROUGE-LSum (f-measure) with the default stemmer on.\footnote{Using pypi package \texttt{rouge-score}.}
    \item \textbf{Disambig-F1} (Disambiguated F1 Accuracy): Measures the correctness of the system-generated answer. Given the gold-standard disambiguated QA pairs, we run a reading comprehension model (RoBERTa \citealp{liu2019roberta} trained on SQuAD 2.0 \citealp{rajpurkar-etal-2018-know}) where the system-generated long-form answer is the context. We then evaluate the number of disambiguated questions that can be answered using the long-form answer as context by calculating the Q1-F1 accuracy.
    \item \textbf{DR} (Disambiguation-ROUGE): The geometric mean of ROUGE-L and Disambig-F1 which penalizes methods that maximize one metric over another.
\end{itemize}
Finally, we also report the average number of words of the system-generated answers.

\subsection{Results}

We compared several systems which can be divided into three types: {\em few-shot prompting}, {\em prompt tuning}, and {\em finetuning}.

For few-shot prompting, we used five exemplars in the prompt, and compared several
prompt configurations using the 540B-sized PaLM model \citep{chowdhery2022palm}:
\begin{itemize}
    \item Random exemplars (random): Select at random five QA pairs from the training dataset and use as exemplars. We report the average results of five different runs.
    \item Query diversified exemplars (QD): Select five QA pairs with questions that have different classes of ambiguity as classified in Section~\ref{sec:qtypes}. We only considered the 100 exemplars with questions labeled as one of five multifacetedness types, which are originally sampled from the ASQA training dataset (i.e., top 5 in Table~\ref{tab:qtypes}). 
    \item NL/QA/AF refinement: Include a query refinement step and dynamically select exemplars from a pool of exemplars.
\end{itemize}
The first block of Table~\ref{tab:asqa} reports the results from few-shot prompting. As can be seen, using exemplars with questions of different types of ambiguity significantly improves over the random baseline. All query refinement prompts improve the performance, where the AF refinement performs the best among them. 

\begin{table}[t]
    \footnotesize
    \centering
    \begin{tabular}{lcccc}
        \toprule
        \textbf{Model} & \textbf{\#Words} & \textbf{ROUGE-L} & \textbf{Disambig-F1} & \textbf{DR} \\
        \midrule
        \multicolumn{5}{c}{\textit{Few-shot Prompting}} \\
        \midrule
        PaLM 540B (random) & 75.5 & 31.1 & 18.6 & 24.0 \\
        PaLM 540B (QD) & 62.1 & 31.3 & 22.8 & 26.7 \\
        \quad + NL refinement & 63.6 & 33.6 & 23.9 & 28.3 \\
        \quad + QA refinement & 41.9 & 32.3 & \textbf{25.3} & 28.6 \\
        \quad + AF refinement & 62.4 & \textbf{34.5} & \textbf{25.3} & \textbf{29.6} \\
        \midrule 
        GPT 175B (random)${}^*$ & 21.6 & 12.9 & 7.7 & 10.0 \\
        \quad + AF refinement & 46.5 & 30.0 & 17.1 & 22.6 \\
        InstructGPT-3 175B (random) & 40.7 & 31.8 & 25.0 & 28.2 \\
        \quad + AF refinement & 39.0 & 31.6 & 23.4 & 27.2 \\
        \midrule
        \multicolumn{5}{c}{\textit{Few-shot Prompt Tuning}} \\
        \midrule
        PaLM 540B 100-shot & 62.5 & 36.1 & 25.0 & 30.0 \\
        \quad + AF refinement & 53.8 & \textbf{36.7} & \textbf{25.4} & \textbf{30.6} \\
        \midrule
        \multicolumn{5}{c}{\textit{Using the full dataset}} \\
        \midrule
        T5-Large Closed Book & 62.5 & 31.0  & 7.4 &  15.1 \\
        T5-Large Open Book 1 Passage & 63.0 & 36.5 & 21.2 & 27.9 \\
        T5-Large Open Book 3 Passages & 71.1 & 38.8& 25.1 & 31.2 \\
        T5-Large Open Book 5 Passages & {71.6} & \textbf{39.2} & {26.4} & \textbf{32.1}\\
        PaLM 540B Prompt Tuning & 64.1 & 37.4 & \textbf{27.8} & \textbf{32.1} \\
        \bottomrule
    \end{tabular}
    \caption{Evaluation of several systems on the dev set of the ASQA dataset. The best values for each setting are \textbf{bold}-faced. We mark systems with an asterisk (${}^*$) if they failed to generate answers for at least half the total number of examples.}
    \label{tab:asqa}
\end{table}

To check whether our prompts work with other models, we also test our prompts with 175B-sized GPT-3 \citep{brown2020language} and InstructGPT-3 \citep{ouyang2022training} models, named \texttt{davinci} and \texttt{text-davinci-002}, respectively, the latter finetuned further with humans in the loop to better follow user instructions.\footnote{
We note that the training data used for \texttt{text-davinci-002} is unknown, i.e., we do not know whether the model had access to supervision from long-form question answering datasets during its training.
} As shown in the second block of Table~\ref{tab:asqa}, using standard prompts in GPT-3 fails to generate answers for at least half the total number of examples. Just by adding our proposed configurations, the performance of GPT-3 significantly increases. Interestingly, we do not see the same increasing trend with the InstructGPT-3 model.

Our prompts can also be applied to prompt tuning \citep{lester-etal-2021-power} where a set of learned embeddings called \textit{soft prompts} is prepended to the prompt. In particular, we follow the method in \citet{rubin-etal-2022-learning}, i.e., we prepend one soft prompt\footnote{
We tried several soft prompt lengths and found that increasing the length beyond one prompt does not lead to any improvements.
} to the input and finetune it using the 100 exemplars we used for dynamic prompting.
The third block in Table~\ref{tab:asqa} reports prompt tuning results, where we see a slight improvement when applying our AF refinement prompts.

Finally, we compare our few-shot systems with systems that are trained with the full dataset.
In the final block of Table~\ref{tab:asqa}, we show results reported in \citet{stelmakh2022asqa}, which are T5-large models in both closed-book (no retrieved passages) and open-book (1/3/5 retrieved passages using Joint Passage Ranker; \citealp{min-etal-2021-joint}) settings. We also report PaLM 540B prompt tuned using the full dataset.
Our best few-shot systems are surprisingly competitive compared to fully finetuned T5 systems, outperforming the closed book system and the open book system with one retrieved passage. Moreover, the correctness of our best systems as measured by the Disambig-F1 score is on par with the open book T5 models.
Finally, prompt tuning PaLM using the full dataset performs the best among all systems in terms of correctness, despite having a lower ROUGE-L score than the best T5 system.
We believe that open-book models have higher ROUGE-L scores due to
the fact that they have access to retrieved passages that they can directly copy, and which may follow the format of the human-generated answers. We discuss this and other annotator biases further in Section~\ref{sec:asqa-biases}.

\subsection{Ablation Studies}

We conducted ablation studies on the best few-shot prompting configuration, which is shown in Table~\ref{tab:ablation}.
In terms of the number of exemplars, while increasing it from 1 to 5 improves the performance, increasing it from 5 to 10 slightly decreases the overall performance. 
However we see an increase in ROUGE-L and STR-EM, which shows that access to more data increases the ability of the model to copy answer formats.
Moreover, increasing the number of dynamic exemplars leads to performance improvements, which is unsurprising.
Also, when removing one component (refinement or dynamic prompting), we see a substantial decrease in performance, where the decrease is larger when refinement is not used.
Using a different similarity metric for prompt selection does not have significant changes in the model performance, however model-based metrics such as BERTScore \citep{Zhang*2020BERTScore:} and BLEURT \citep{sellam-etal-2020-bleurt} are slightly better than string-based metrics such as BM25 \citep{robertson1995okapi}.
Finally, the increase in performance of using AF query refinement prompts can also be seen when using smaller versions of PaLM.

\begin{table}[t]
    \footnotesize
    \centering
    \begin{tabular}{lccccc}
        \toprule
         \textbf{Model} & \textbf{\#Words} & \textbf{ROUGE-L} & \textbf{Disambig-F1} & \textbf{DR} \\
        \midrule
        \multicolumn{5}{c}{\textit{Number of exemplars in the prompt}} \\
        \midrule
        1-shot & 55.1 & 31.7 & 23.2 & 27.1 \\
        3-shot & 37.9 & 32.0 & 23.7 & 27.6 \\
        {5-shot} & 62.4 & 34.5 & 25.3 & 29.6 \\
        10-shot & 66.0 & 34.9 & 24.6 & 29.3  \\
        \midrule
        \multicolumn{5}{c}{\textit{Number of dynamic prompt exemplars}} \\
        \midrule
        5 & 69.0 & 35.0 & 22.1 & 27.8 \\
        25 & 68.0 & 33.9 & 23.2 & 28.0 \\
        50 & 63.8 & 34.6 & 23.6 & 28.6 \\
        {100} & 62.4 & 34.5 & 25.3 & 29.6\\
        \midrule
        \multicolumn{5}{c}{\textit{Without one component}} \\
        \midrule
        Without refinement & 66.3 & 33.2 & 22.9 & 27.6 \\
        Without dynamic prompting & 40.1 & 32.5 & 25.1 & 28.6 \\
        \midrule
        \multicolumn{5}{c}{\textit{Similarity metric for prompt selection}} \\
        \midrule
        {BERTScore}  & 62.4 & 34.5 & 25.3 & 29.6 \\
        BLEURT & 65.3 & 34.6 & 25.1 & 29.5 \\
        BM25 & 64.4 & 34.6 & 24.4 & 29.0 \\
        \midrule
        \multicolumn{5}{c}{\textit{Smaller models}} \\
        \midrule
        PaLM 8B (random) & 84.1 & 21.1 & 9.2 & 14.0 \\
        \quad + AF refinement & 63.9 & 29.6 & 10.0 & 17.2\\
        PaLM 62B (random) & 66.6 & 28.7 & 14.5 & 20.4\\
        \quad + AF refinement & 65.3 & 32.4 & 18.0 & 24.1 \\
        \bottomrule
    \end{tabular}
    \caption{Performance of ablated versions of the best few-shot prompting configuration.}
    \label{tab:ablation}
\end{table}

\subsection{Further Analyses}
\label{sec:asqa-biases}

In the next paragraphs, we discuss annotator biases in ASQA that few-shot systems would not be able to capture.

\paragraph{Question Disambiguation Analysis}
The first bias is in the way questions are disambiguated. We observed cases where the majority ambiguity class selected from dynamic prompting was a plausible type of ambiguity for the given question, but the particular disambiguation in ASQA is of another type. This leads to few-shot systems generating an entirely different yet also correct answer to the question, an example of which is shown in the first block of Table~\ref{tab:examples}. The gold and the T5 answers disambiguated the question based on the habitat of the animal (fresh water or saltwater), while the PaLM answer disambiguated it based on both the habitat and the characteristics (having gills, having lungs, or having both).
In fact, by just providing how the ASQA dataset expects the question to be disambiguated (prompt template shown in Section~\ref{sec:prompt-templates}), the performance of our best configuration improves by up to 8.8 DR points (29.6 vs 38.4).
Future work on long-form question answering evaluation should explore methods to deal with questions that can be disambiguated in multiple ways.

\paragraph{Summary Format Analysis}
The second bias is the way long-form answers are annotated. The human-written long-form answers from all the splits of ASQA are written by the same set of annotators using the same annotation template. This creates a formatting bias that models finetuned using thousands of examples can capture. We attribute the low ROUGE-L scores of few-shot systems, in comparison to finetuned systems, to this bias.
The second example in Table~\ref{tab:examples} shows this bias, where both answers are equally answer the question, but are written differently and thus have a significant difference in ROUGE-L. While both systems correctly answered \textit{Tony Hawk} and \textit{Danny Way}, the T5 long answer looks more similar to the gold-standard answer, whereas the PaLM long answer looks more concise and arguably more readable.
This finding is related to that of \citet{goyal2022news}, where they showed that document summaries generated by GPT-3 are always preferred by humans but are ranked the lowest by ROUGE.
Future work on evaluation should explore reference-free evaluation, where answers are instead compared to trustworthy sources, similar to \citet{rashkin2021measuring}.

\begin{table}[t]
    \centering
    \footnotesize
    \begin{tabular}{p{0.95\textwidth}}
        \toprule
        {\textit{Question}}: What do you call animals live in water? \\
        \textit{Gold}: \textbf{Marine life, or sea life or ocean life}, is the plants, animals and other organisms that live in the salt water of the sea or ocean, or the brackish water of coastal estuaries. An \textbf{aquatic animal} is an animal, either vertebrate or invertebrate, which lives in the water for most or all of its lifetime. The term aquatic can be applied to animals that live in either fresh water or salt water. However, the adjective marine is most commonly used for animals that live in saltwater, i.e. in oceans, seas, etc. \\ \\
        \textit{T5 best}: \textbf{aquatic animals} are animals that live in water for most or all of their lifetime. aquatic animals can be either vertebrate or invertebrate. fresh water animals are \textbf{aquatic animals}. \textbf{marine animals} are animals that live in saltwater. \\
        \multicolumn{1}{r}{\mbox{\textbf{ROUGE-L}:~44.96}}  \\ 
        \textit{PaLM best}: Animals that live in water are called \textbf{aquatic animals}. Aquatic animals that have gills are called \textbf{fish}. Aquatic animals that have lungs are called \textbf{marine mammals}. Aquatic animals that have both gills and lungs are called \textbf{amphibians}.\\
        \multicolumn{1}{r}{\mbox{\textbf{ROUGE-L}:~20.16}}  \\ 
        \midrule
        {\textit{Question}}: Who landed the first 900 on a skateboard? \\
        \textit{Gold}: The 900 is a 2 1/2-revolution (900 degrees) aerial spin performed on a skateboard ramp. A number of opinions exist about pre-1999 900s. The most prominent of these is the argument that \textbf{Danny Way} landed the 900 in 1989, and it appears in an early Santa Cruz film. \textbf{Tony Hawk}, one of the most successful vertical pro skateboarders in the world, is generally said to have landed the first 900 on a skateboard. He landed the 900 at X Games V in 1999 after ten failed attempts. \\ \\
        \textit{T5 best}: the 900 is a 2 1/2-revolution (900 degrees) aerial spin performed on a skateboard ramp. a number of opinions exist about pre-1999 900s. the most prominent of these is the argument that \textbf{danny way} landed the 900 in 1989, and it appears in an early santa cruz film. in 1999 \textbf{tony hawk}, one of the most successful vertical pro skateboarders in the world, landed "the 900" at the 1999 x-games after ten failed attempts. in october 2011, schaar became the 8th person in history to land a 900 on a skateboard. in september 2016, \textbf{schaar} became the 1st person in history to land a stalefish 900 on a skateboard. \\
        \multicolumn{1}{r}{\mbox{\textbf{ROUGE-L}:~80.00}}  \\ 
        \textit{PaLM best}: The 900 is a skateboarding trick in which the skateboarder does two and a half vertical revolutions. American professional skateboarder \textbf{Tony Hawk} landed the first 900 in a competition at the X Games in 1999. Australian professional skateboarder \textbf{Danny Way} landed the first 900 on video in 1990.\\  
        \multicolumn{1}{r}{\mbox{\textbf{ROUGE-L}:~45.59}} \\
        \bottomrule
    \end{tabular}
    \caption{Example ASQA outputs from the T5 and PaLM best configurations. Disambiguated answers are \textbf{bold}-faced.}
    \label{tab:examples}
\end{table}

\section{Experiments on Query-focused Multi-document Summarization}
\label{sec:results-aquamuse}

\subsection{Dataset}

The AQuAMuSe dataset \citep{kulkarni2020aquamuse} is a query-focused multi-document summarization dataset, which was created to simulate how a search engine would consolidate information from multiple documents of high relevance to a given query. 
The dataset is also a subset of the NQ dataset \citep{kwiatkowski-etal-2019-natural}, but is extended with web documents extracted from Common Crawl and long-form answers from Wikipedia.
The dataset consists of 6,599, 714, and 849 training, development, and test examples, where each example is given, on average, 6.46 web documents (2,008 tokens per document). We note that in the closed book setting, the web documents are not used.


\subsection{Evaluation}

We compare systems using an n-gram overlap-based metric ROUGE-1/2/L \citep{lin-2004-rouge} and a QA-based metric QAEval${}_\text{rheme}$ \citep{narayan2022conditional}. QAEval${}_\text{rheme}$ is similar to the original QAEval \citep{deutsch-etal-2021-towards}, where a question generation model is used to generate questions from the gold summary, and a question answering model attempts to answer these questions using the system-generated summary as the context.
In QAEval${}_\text{rheme}$, the way questions are generated is modified such that we only consider questions that are information-seeking (i.e., based on the theme-rheme structure; \citealp{vallduvi1998rheme,kruijff2003discourse}).

\subsection{Results}

We compared the following systems. Few-shot prompting systems include PaLM 540B using random exemplars, and using NL/QA/AF refinements. 
For the refinements, we used the 20 exemplars labeled with the {\em Needs Elaboration} type, which are sampled from the AQuAMuSe training dataset (i.e., the sixth type in Table~\ref{tab:qtypes}).  
We also experimented including from ASQA prompt exemplars during dynamic prompt selection to allow the model to more effectively differentiate question types. Few-shot prompt tuning systems include PaLM 540B, with and without AF refinement, and with ASQA prompt exemplars during dynamic prompt selection.
Finally, fully finetuned systems include T5-XL closed book and open book, LongT5-XL \citep{guo-etal-2022-longt5} open book that allows longer contexts, and PaLM 540B prompt tuning. For the open book systems, we filled in the context with as many passages as possible within the limits of their input lengths. This results to T5 having 8.3 passages and LongT5 having 30.0 passages on average.

Table \ref{tab:aquamuse} reports the scores of these systems. Here, we see similar results as in ASQA; in the few-shot prompting and prompt tuning systems, the addition of AF refinements improves the most over the random baseline among the three types of refinements. Moreover, the inclusion of ASQA prompt exemplars during dynamic prompt selection also substantially improves the scores.
It is interesting to note that, in contrast to ASQA results, prompt tuning performs worse than few-shot prompting, having worse QAEval${}_\text{rheme}$ scores overall.
When compared with finetuning systems, our prompting systems significantly outperform the closed book variant of T5, but fall behind the open book systems. 
We believe that this is due to the fact that AQuAMuSe summaries are extractive -- only 2\%/13\%/24\%/31\% of the unigrams/bigrams/trigrams/4-grams in the summaries are novel. This allows open-book systems to just copy directly from the source.



\begin{table}[t]
    \footnotesize
    \centering
    \begin{tabular}{lcccc}
        \toprule
        System & ROUGE-1 & ROUGE-2 & ROUGE-L & QAEval${}_\text{rheme}$ \\
        \midrule
        \multicolumn{5}{c}{\textit{Few-shot Prompting}} \\
        \midrule
        PaLM 540B (random) & 31.52 & 15.66 & 28.12 & 10.26 \\
        \quad + NL refinement & 37.52 & 17.34 & 33.62 & 12.20 \\
        \quad + QA refinement & 34.78 & 16.97 & 31.21 & 11.50\\
        \quad + AF refinement & 36.84 & 16.92 & 32.94 & 12.50 \\
        \quad \quad \quad + ASQA exemplars & \textbf{37.72} & \textbf{18.11} & \textbf{33.73} & \textbf{13.24}\\
        \midrule
        \multicolumn{5}{c}{\textit{Few-shot Prompt Tuning}} \\
        \midrule
        PaLM 540B 20-shot & 34.12 & 11.89 & 29.63 & 7.96\\
        \quad + AF refinement & 34.33 & 12.43 & 30.03 & 9.13 \\
        \quad \quad \quad + ASQA exemplars & \textbf{37.45} & \textbf{16.55} & \textbf{33.30} & \textbf{11.77} \\
        \midrule
        \multicolumn{5}{c}{\textit{Using the full dataset}} \\
        \midrule
        T5-XL Closed Book & 29.39 & 10.45 & 26.17 & 5.62 \\
        T5-XL Open Book 8.3 Passages & 44.93 & 27.10 & 41.53 & 19.77 \\
        LongT5-XL Open Book 30.0 Passages & \textbf{64.11} & \textbf{50.60} & \textbf{61.43} & \textbf{39.22} \\
        PaLM 540B Prompt Tuning & 40.53 & 19.66 & 36.56 & 14.62 \\
        \bottomrule
    \end{tabular}
    \caption{Evaluation of several systems on the test set of the AQuAMuSe dataset. The best values for each block are \textbf{bold}-faced.}
    \label{tab:aquamuse}
\end{table}

\section{Conclusions}

In this paper, we investigated the ability of large language models to answer questions in a long-form manner through two question answering datasets: ASQA and AQuAMuSe. We introduced query refinement prompting that improves over standard few-shot prompting and prompt tuning methods by encouraging the model to explicitly express the multifacetedness in questions. 
With the use of query refinement prompts on both few-shot closed book prompting and prompt tuning settings, we are able to outperform systems trained using the full training data in both ASQA and AQuAMuSe. While we also achieve comparable results with open book systems in ASQA, we acknowledge that these systems still perform better when they have access to more retrieved passages. For future work, we plan to explore ways to few-shot prompt large language models in the open book setting and ways to augment a retrieval component into large language models.

\section*{Acknowledgements}

The authors would like to thank Tal Schuster and Mirella Lapata for their detailed feedback, and Ji Ma and Priyanka Agrawal for their help in prompt tuning experiments. Finally, we are thankful to the PaLM team in Google Research for providing easy-to-use tools to experiment with large language models.


\bibliography{anthology,custom}
\bibliographystyle{ICLR23/iclr2023_conference}

\appendix

\section{Prompt Templates}
\label{sec:prompt-templates}

This section presents the prompt templates we used in our experiments. Figures~\ref{fig:asqa-prompt-template}~\ref{fig:aquamuse-prompt-template} show prompt templates with AF refinements for ASQA and AQuAMuSe, respectively. 
When using no refinements, the \texttt{Disambiguations} part of the template is removed. When using a different query refinement step, the same part is replaced with a natural language explanation or a list of question answers.
Finally, Figure~\ref{fig:disambigq-template} shows the template used when the oracle disambiguated questions are given.

\begin{figure}
    \centering
    \footnotesize
    \begin{tabular}{p{0.95\textwidth}}
    \toprule
    \texttt{I will provide ambiguous questions that have multiple answers about different aspects of the question, and answer them in detail with at least two sentences.} \\
    {} \\
    \texttt{Question: Who sang it's a long way to the top?} \\
    \texttt{Disambiguations:} \\
    \texttt{- band: AC/DC} \\
    \texttt{- lead vocal: Bon Scott} \\
    \texttt{Answer: "It's a Long Way to the Top (If You Wanna Rock 'n' Roll)" is a song by Australian hard rock band AC/DC. This was a signature song for lead singer Bon Scott. Brian Johnson, who replaced Scott as AC/DC's lead vocalist after Scott's death in 1980, does not perform it, out of respect for his predecessor.} \\
    {} \\
    \texttt{Question:} \\
    \bottomrule
    \end{tabular}
    \caption{Example ASQA Prompt with AF refinements and one exemplar.}
    \label{fig:asqa-prompt-template}
\end{figure}

\begin{figure}
    \centering
    \footnotesize
    \begin{tabular}{p{0.95\textwidth}}
    \toprule
    \texttt{I will provide questions that need to be elaborated 
                  to be answered fully, and will answer them in detail with
                  at least two sentences.} \\
    {} \\
    \texttt{Question: where did the term shooting brake come from} \\
    \texttt{Details:} \\
    \texttt{- how the term originated: as an early 19th century British term} \\
    \texttt{- what it was for: a vehicle used to carry shooting parties with their equipment and game} \\
    \texttt{- etymology of the term brake: uncertain; initially a chassis used to break in horses, used to describe a motorized vehicle} \\
    \texttt{- its possible origins: in the Dutch word 'brik' which means 'cart' or 'carriage'"} \\
    \texttt{Answer: "Shooting-brake" originated as an early 19th century British term for a vehicle used to carry shooting parties with their equipment and game. The etymology of the term brake is uncertain; initially a chassis used to break in horses, and subsequently used to describe a motorized vehicle. It is also possible, that the word' brake' has its origins in the Dutch word' brik' which means' cart' or' carriage'.} \\
    {} \\
    \texttt{Question:} \\
    \bottomrule
    \end{tabular}
    \caption{Example AQuAMuSe Prompt with AF refinements and one exemplar..}
    \label{fig:aquamuse-prompt-template}
\end{figure}

\begin{figure}
    \centering
    \footnotesize
    \begin{tabular}{p{0.95\textwidth}}
    \toprule
    \texttt{I will provide ambiguous questions that have multiple answers about different aspects of the question, and answer them in detail with at least two sentences.} \\
    {} \\
    \texttt{Question: Who sang it's a long way to the top?} \\
    \texttt{Disambiguated Questions:} \\
    \texttt{Q: Which band sang it's a long way to the top?}  \\
    \texttt{Q: Who was the lead vocal of it's a long way to the top?}  \\
    \texttt{Disambiguated Answers:} \\
    \texttt{- band: AC/DC} \\
    \texttt{- lead vocal: Bon Scott} \\
    \texttt{Answer: "It's a Long Way to the Top (If You Wanna Rock 'n' Roll)" is a song by Australian hard rock band AC/DC. This was a signature song for lead singer Bon Scott. Brian Johnson, who replaced Scott as AC/DC's lead vocalist after Scott's death in 1980, does not perform it, out of respect for his predecessor.} \\
    {} \\
    \texttt{Question:} \\
    \bottomrule
    \end{tabular}
    \caption{Example ASQA Prompt with AF refinements, oracle disambiguated questions, and one examplar.}
    \label{fig:disambigq-template}
\end{figure}



\end{document}